# SKIN DETECTION OF ANIMATION CHARACTERS

Kazi Tanvir Ahmed Siddiqui[1] and Abu Wasif[2]

[1]Department of Electrical and Computer Engineering, North South University, Dhaka, Bangladesh
tanvir.mclv@gmail.com
[2]Department of Computer Science and Engineering, Bangladesh University of Engineering and Technology, Dhaka, Bangladesh
wasif@cse.buet.ac.bd

## ABSTRACT

*The increasing popularity of animes makes it vulnerable to unwanted usages like copyright violations and pornography. That's why, we need to develop a method to detect and recognize animation characters. Skin detection is one of the most important steps in this way. Though there are some methods to detect human skin color, but those methods do not work properly for anime characters. Anime skin varies greatly from human skin in color, texture, tone and in different kinds of lighting. They also varies greatly among themselves. Moreover, many other things (for example leather, shirt, hair etc.), which are not skin, can have color similar to skin. In this paper, we have proposed three methods that can identify an anime character's skin more successfully as compared with Kovac, Swift, Saleh and Osman methods, which are primarily designed for human skin detection. Our methods are based on RGB values and their comparative relations.*

## 1. INTRODUCTION

Skin identification from an image is the precondition of various applications such as: face detection and recognition, obscene video recognition, people tracking, hand gesture recognition and mood identification. Face detection/recognition is used in almost everywhere, for search engines, digital cameras and surveillance systems. But, these techniques are made for humans only. Most of the cases, these methods do not work on animes, cartoons or on comic characters.

Characters in cartoons, comics and games are collectively called cartoon characters [1]. Currently google image search can detect human faces, but are unable to detect animation faces [2]. Many different techniques are used for skin detection modelling [3-8]. Simpler of those models are based on several threshold values [5-8], while more complex models use Neural Networks [9], Bayesian classifier [10], max entropy [11], k-means clustering [12].

Skin color varies greatly not only for humans but also for animation characters. So, it is very challenging to set a rigorous condition in which skin of all animation characters fit. Skin color differs greatly in different illumination conditions. Another challenge is that, many objects have also skin like color, e.g. wood, leather, hair, sand, skin colored clothing etc. [8]. Even animation characters have different skin colors based on race, gender and age.

In this paper we have proposed three simple methods based on RGB range and their comparative relations. We have used means, standard deviations and ranges from the skin of the cartoon characters to specify the conditions. We have compared our results with other human-skin detection methods [5-8] and an animation-skin detection method [1] and presented the results.

The database we have used is also compiled by us. We have used the google search engine, and an online database [13] to accumulate the images used in our experiment. We have also annotated skins manually using Adobe Photoshop to set the accurate skin pixels.

## 2. RELATED WORKS

### 2.1. Human Skin Detection Methods

There has been several methods for human skin detection. Some popular methods are: Kovac et al. [5], Saleh [6], Swift [7], Osman et al. [8], Omanovic et al. [14] etc. These all methods are based on specific RGB range, RGB ratios and their comparative relations. These methods are described below:

According to Kovac method, a pixel **is skin** if:

- R > 95 and G > 40 and B > 20 **and**
- Max (R, G, B) –Min(R, G, B) > 15 **and**
- |R – G| > 15 **and**
- R > G and R > B

According to Swift method, a pixel is **not skin** pixel if:

- B > R **or**
- G < B **or**
- G > R **or**
- B < R/4 **or**
- B > 200

According to Saleh method, a pixel **is skin** if:

- R – G > 20 **and**
- R – G < 80

Osman et al. [8] and Omanovic et al. [14] methods are based on RGB ratio.

According to Osman method, a pixel **is skin** if:

- $0.0 \leq (R-G)/(R+G) \leq 0.5$ **and**
- $B/(R+G) \leq 0.5$

Omanovic et al. [14] tested with various RGB ratios and showed their performance accordingly.

### 2.2. Cartoon character skin detection method

Anime-skins are not identical to human skins. They are drawn by artists and their color varies greatly based on age, race, gender, illumination etc. Takayama et al. [1] presented a way of detecting anime-skin. Their method uses HSV values.

According to Takayama et al. [1], a pixel **is skin** if:

- Hue is between **0 to 40** degrees **and**
- Value is more than **75%**

This method uses Canny method [15] for edge extraction, and then, the skin region segmentation is done by the flood-fill approach.

## 3. METHODOLOGY

The purpose of this paper is to establish a method for skin extraction from animation or cartoon images. The total task can be divided into four steps:

1. Data collection
2. Data preparation
3. RGB based skin detection methods
4. Testing and Evaluation

### 3.1. Data Collection

There are no suitable databases for animation-images. So, we created our own database, which consists of 255 jpeg. images of anime characters. The database is compiled by images collected from an anime database [13] and google image search [2]. Some properties of our database are:

- All the images are color images, no grayscale or binary images were used.
- Images are of both genders.
- Images are of different illumination conditions e.g. images were taken at morning, dusk and night.
- Image characters are from different races e.g. White, Black and Asian.
- Backgrounds of the images are cluttered and also of uniform color.
- Some images consist of multiple characters while others single characters.
- Characters in the images have varied hair color: black, white, brown, grey, red etc.
- Some subjects of the images have spectacles, headbands, armbands and tattoos.

Image size ranged from dimensions 104×99 to 1003×1200. We have used 70 images from the 255 images to set the conditions of our methods.

### 3.2. Data Preparation

An accurate image segmentation is one of the most important steps to establish the ground truth information and to get the most accurate result [4]. In order to annotate skin from image, human intervention is required. We have annotated 70 images manually using Adobe Photoshop CS6. We have used the magic wand tool of that software to manually select skin areas. It is very difficult, even for humans, to identify whether some pixels are skin or not, because of the different illumination conditions and shadows. Moreover, our images are of anime characters. These images were drawn by humans. Hence, we have established some rules to maintain integrity while annotating images:

- Edges were **not considered as skin.**
- Shaded skin, were **considered as skin.**
- Where cloth or hair meet skin, those pixels were **considered skin** based on the difference with other skin pixels. If the difference were too much then they were **not counted as skin.**

We have worked on **70** images, a total of **18,000,593** pixels. We colored those skin pixels using RGB values of: **(0, 255, 0)** or **(255, 255, 0).** This specific values were used so that, the correct amount and location of each skin pixel can be obtained. Next, we compared that specific pixel with our method's found skin pixel to ensure that we have identified correctly or not.

We used MATLAB R2014a software to ensure whether our **manually annotated** skin pixels are detected as skin or not. We observed some anomalies (i.e. some manually colored skin pixels were not detected). That is why, considering some variations, to find all of the skin pixels, we have fixed the range of RGB values according to Table 1.

We used two colors: green (0, 255, 0) and yellow (255, 255, 0) to paint the skins, so that while extracting, skin color does not combine with other background colors.

Table 1.  Algorithm for extracting skin pixels from annotated images

| |
|---|
| Input: manually annotated colored images |
| Output: Matrix with skin pixels |
| For each pixel in the image: |
| If (R < 120 **AND** G > 200 **AND** B < 100)  **OR** (R > 200 **AND** G > 200 **AND** B < 100) |
| Then that pixel will be skin  pixel |
| Create an empty matrix of the size of the image |
| Store 1 for every skin pixel in the matrix |

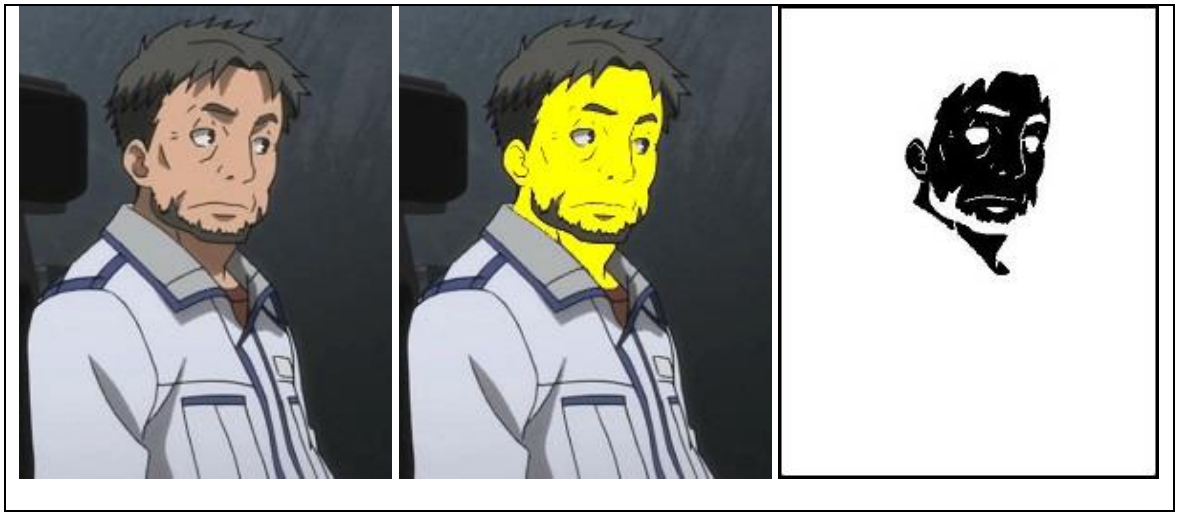

Figure 1. Actual image    Figure 2. Skin colored manually    Figure 3. Manually colored image's skin extracted using Algorithm of Table 1.

Figure 1 shows the actual image before coloring, Figure 2 shows that image colored using Adobe Photoshop (only the skin portion) and Figure 3 shows how the algorithm of Table 1 detects the ground truth of skin pixels from the colored image.

We checked all the images manually to ensure that, every image's skin has been identified correctly.

### 3.3. RGB based skin detection algorithm

In designing our algorithm, we have taken into consideration that skin color varies greatly according to race, gender and illumination conditions. We have collected RGB values from the skin pixels of different images. And then, we used the range, mean and deviations of those values to specify the conditions for detecting skin.

Based on the range and mean of the RGB values of skin and their comparative relations, we have specified some conditions to be skin. One version of the algorithm is described in Table 2

Table 2. Proposed Method-I for Detecting Skin Pixels

Input: Animation images

Output: 2D matrix where 1 is for skin, 0 is for non-skin

For each pixel in the image:

If (120<R<255 **AND** 90<G<250 **AND** 70<B<218 **AND** R>G+10 **AND** G>B+10)

Then that pixel is skin

Create an empty matrix equals the size of the image

Store 1 for every skin pixel

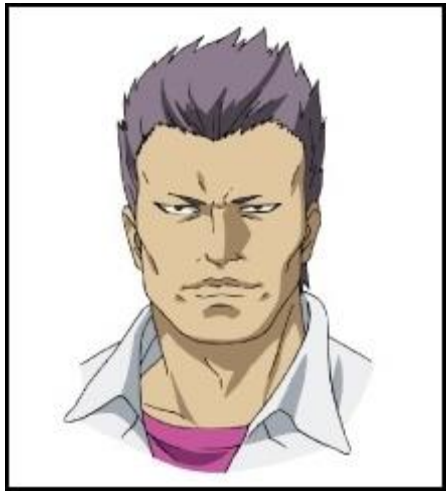 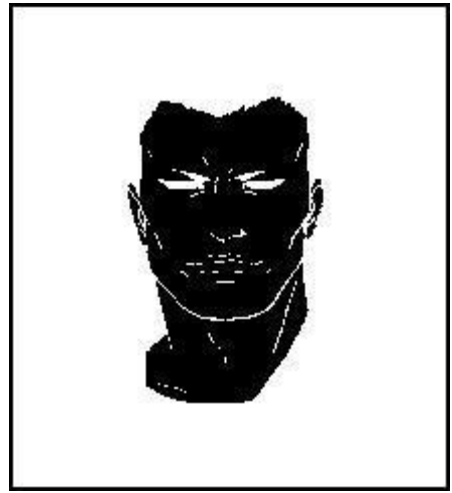

Figure 4.  Actual image          Figure 5. Skin detected using Table 2

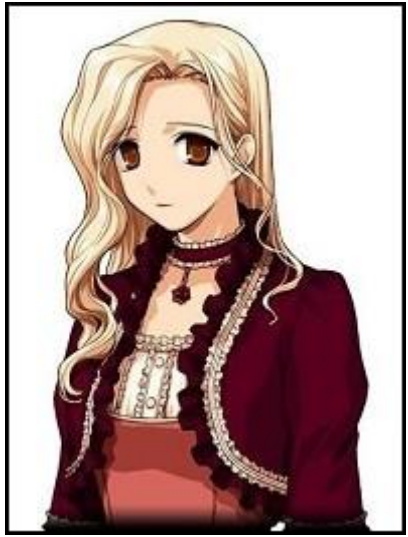 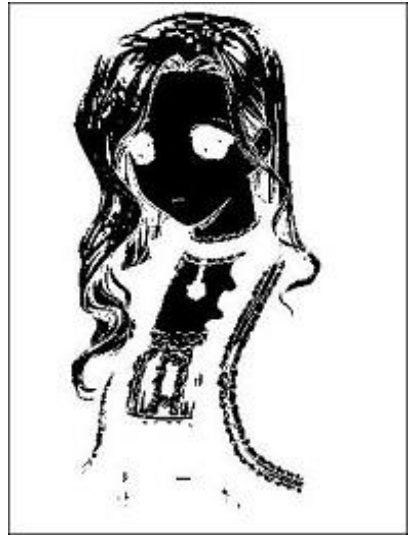

Figure 6.  Actual image    Figure 7.  Skin detected using Table 2

We see that in Figure 5 almost all the skin pixels have been detected. But, in Figure 7, in addition to skin pixels, some hair and dress pixels also have been detected as skin. This is due to the fact that, hair color of this subject is similar to its skin color.

We have changed the conditions slightly to increase the true positive. The changed condition is described in Table 3 and Table 4. These new conditions increase our TP from 79.54% to 83.91% (Table 3) and 88.3% (Table 4) but also increased the FP from 9.5% to 10.98% (Table 3) and 14.2% (Table 4).

Table 3.  Proposed Method-II for Detecting Skin Pixels

Input: Animation images

Output: 2D matrix where 1 is for skin, 0 is for non-skin

For each pixel in the image:

If (120 < R < 255 **AND** 90 < G < 250 **AND** 70 < B < 218 **AND** R > G+10 **AND** G > B)

Then that pixel is skin

Create an empty matrix that equals the size of the image

Store 1 for every skin pixel

Table 4.  Proposed Method-III for Detecting Skin Pixels

Input: Animation images

Output: 2D matrix where 1 is for skin and 0 is for non-skin

For each pixel in the image:
If (120 < R < 255 **AND** 90 < G < 250 **AND** 70 < B < 218 **AND** R > G **AND** G > B)
Then that pixel is skin

Create an empty matrix equals the size of the image

Store 1 for every skin pixel

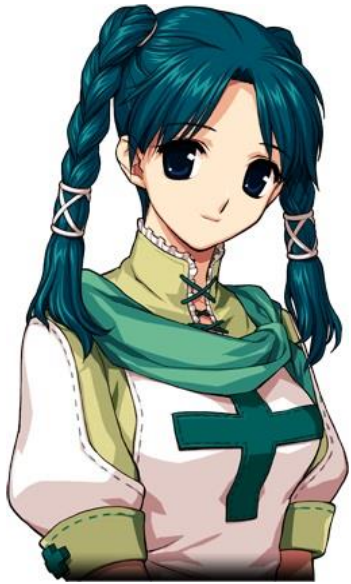 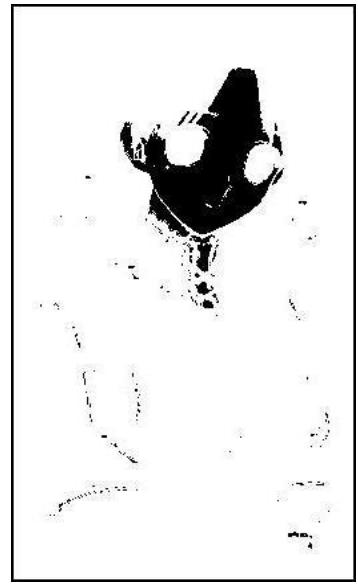

Figure 8.  Actual image　　　　　　　　　Figure 9.  Skin detected using Table 2

Figure 8 shows the actual image. Figure 9, Figure 10 and Figure 11 show distinct differences (shown in red circles). These figures show that, though there is a marked increase in TP; FP is also increased. Figure 12, 13, 14 and 15 show that while Method-I has been failed to detect some of the skin pixels, Method-II and Method III have been successful.

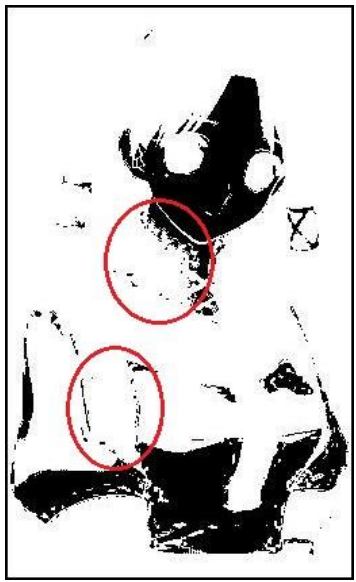 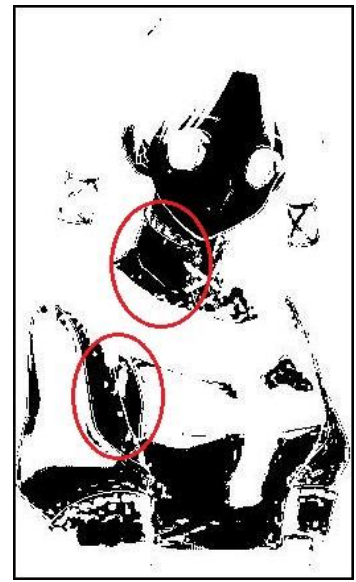

Figure 10.  Skin detected using Table 3　　　Figure 11.  Skin detected using Table 4

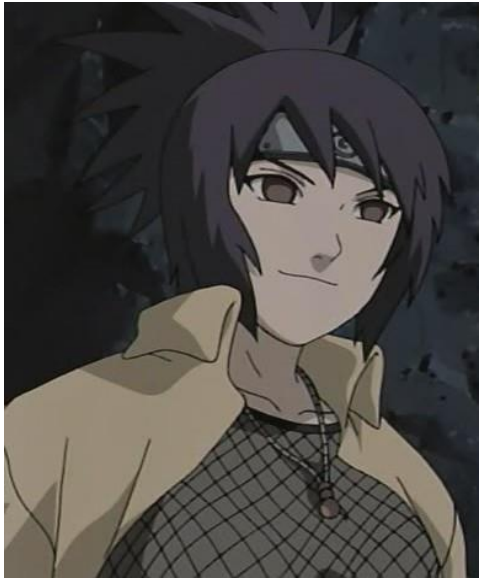
Figure 12. Original image

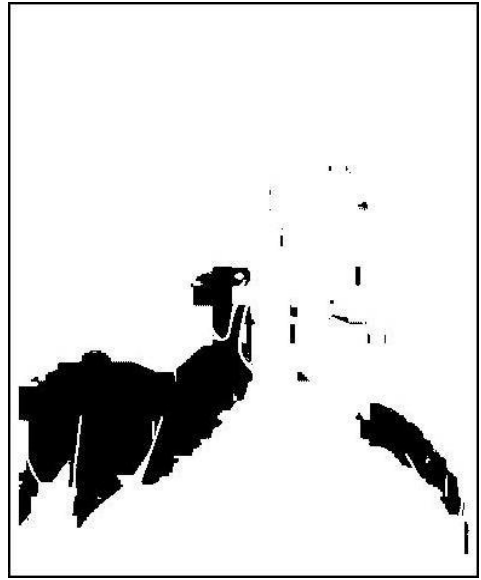
Figure 13. Skin detected using Table 2

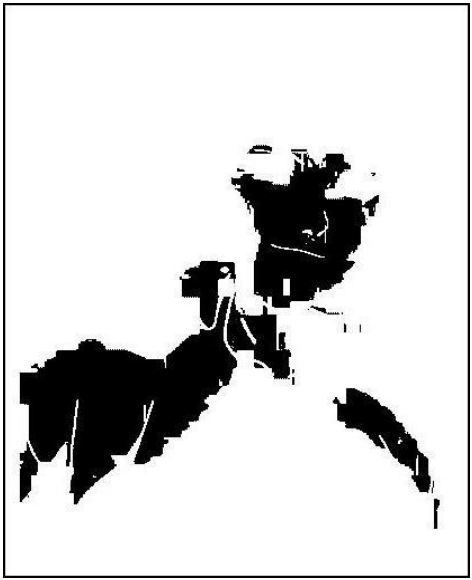
Figure 14. Skin detected using Table 3

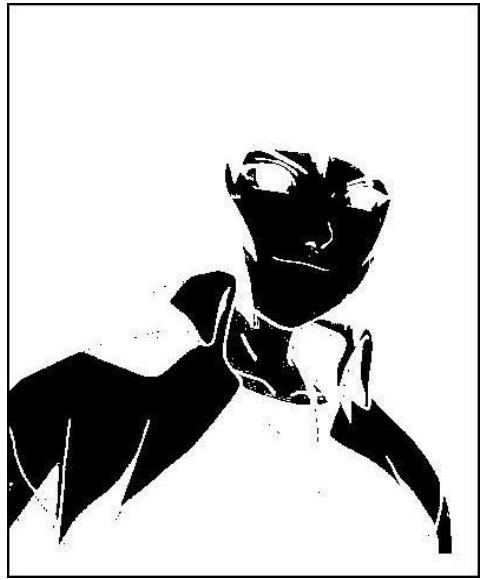
Figure 15. Skin detected using Table 4

We have also used Kovac et al. [5], Saleh [6], Osman et al. [8], and Takayama et al. [1] classifiers to detect skin and compared their performances accordingly.

### 3.4. Testing and Evaluation

Quantitative and qualitative methods are widely used to measure the performance of skin detection algorithm [16]. Quantitative method shows the value of True positive and False positive, while the Qualitative method shows the performance of the method visually. We have used both methods to evaluate our Algorithms performance.

For performance measurement, the true positive (TP) and the false positive (FP) are defined, in the context of our experiment, as:

$$TP = \frac{\text{number of pixels detected as skin}}{\text{total number of skin pixels}}$$

$$FP = \frac{\text{number of non-skin pixels detected as skin}}{\text{total number of pixels in an image}}$$

Table 5. Method to find TP and FP

```
Input: All pixels in the database
Output: TP and FP
For each image in the database:
    Matrix A = Annotated original image
    Matrix B = Image after skin extracted using any method
    For each value of A and B
        IF A and B equals to 1 then TP
        IF B equals to 1 and B > A then FP
    End
End
TP = TP/total number of skin pixel
FP = FP/total number of pixel in an image
```

Table 5 is used to find the TP and FP from the whole database by comparing every image after finding skin with manually annotated image. We have used three methods (Table 1, 2 and 3), alongside Osman et al. [1], Kovac et al. [5], Saleh [6], Swift [7], and Omanovic et al. [14]. We have shown TP and FP of all methods.

## 4. RESULT AND DISCUSSION

Table 6 shows the performance of different methods, i.e. ours, Kovac, Saleh, Osman and Takayama's methods. We have used all those methods on the same dataset and presented the results accordingly. Fig. 16 to Fig. 20 show some examples of qualitative measurement. Here skin pixels are marked as black, and non-skin pixels are white.

Table 6. Performance of Skin Color Classifiers

| Methods | True Positive | Percent(TP) | False Positive | Percent(FP) |
| --- | --- | --- | --- | --- |
| Method-I | 1389666 | 79.54% | 1711682 | 9.5% |
| Kovac | 1504259 | 86.1% | 2644507 | 14.7% |
| Osman | 1716094 | 98.23% | 7794816 | 43.30% |
| Takayama | 1325582 | 75.88% | 1392439 | 7.7% |
| Saleh | 1369588 | 78.40% | 2521295 | 14.01% |
| Method-II | 1465885 | 83.91% | 1976905 | 10.98% |
| Method-III | 1542610 | 88.3% | 2570836 | 14.2% |

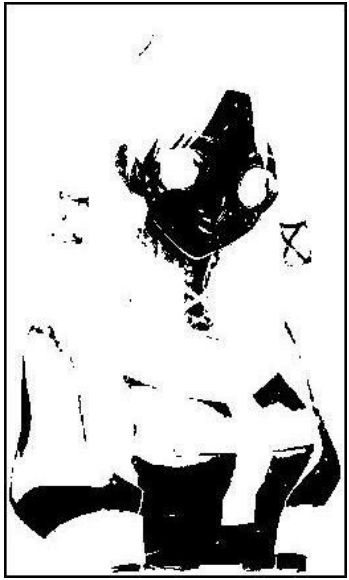
Figure 16. Skin detected using Kovac

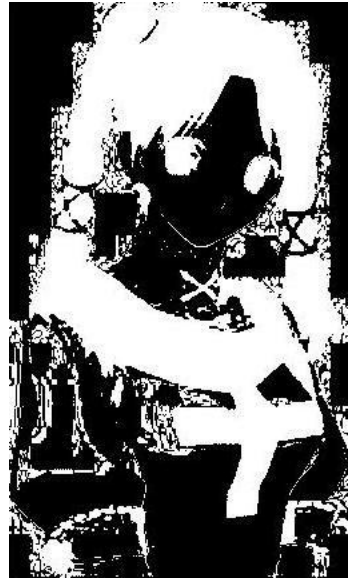
Figure 17. Skin detected using Osman

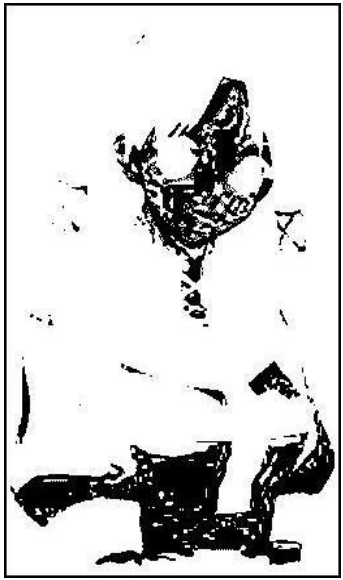
Figure 18. Skin detected using Saleh

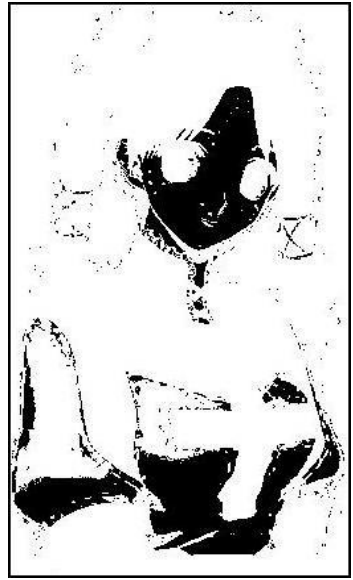
Figure 19. Skin detected using Takayama

Our Method-I extracts the skin of Figure 8, which shows very low FP (Figure 9), while Method-II, III, Kovac, Osman, Saleh and Takayama methods (Figure 10, 11, 16-19) show very high FP.

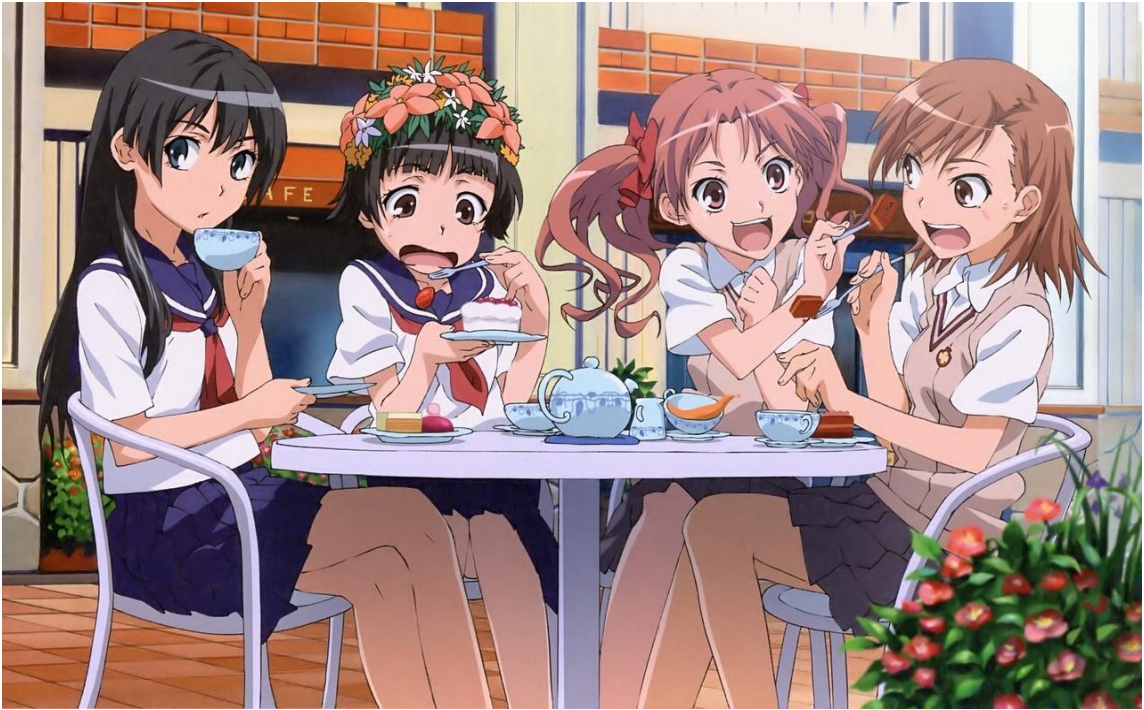

Figure 20. Original Image (multiple characters)

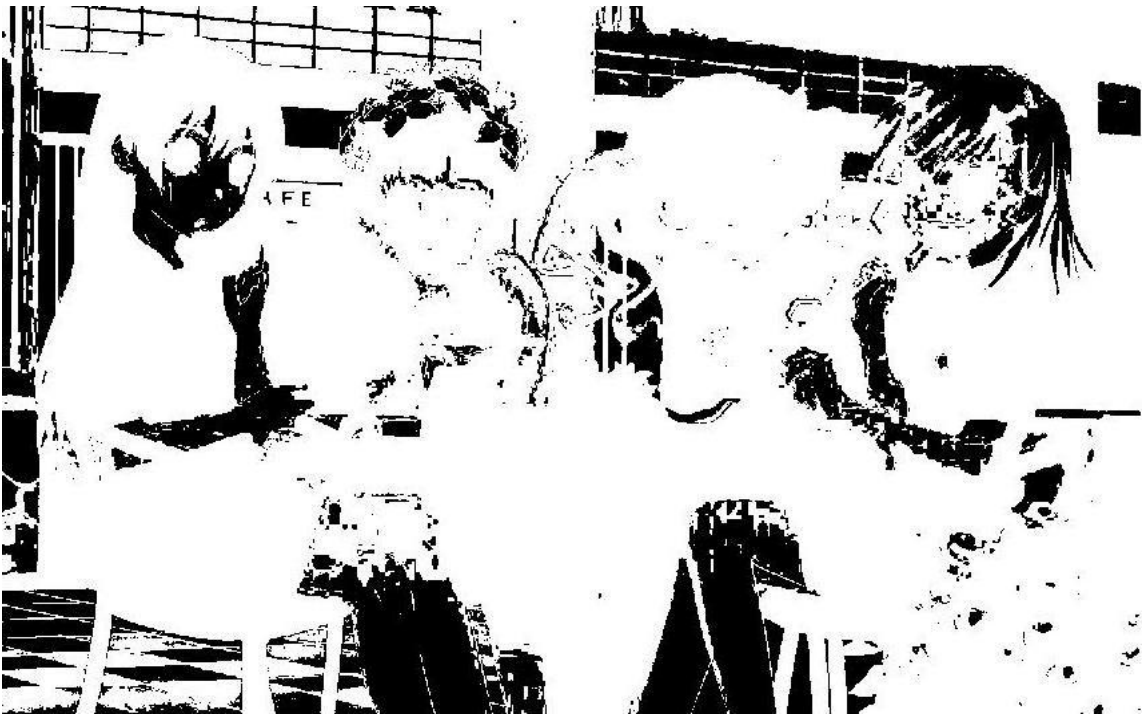

Figure 21. Skin extracted using Method-I

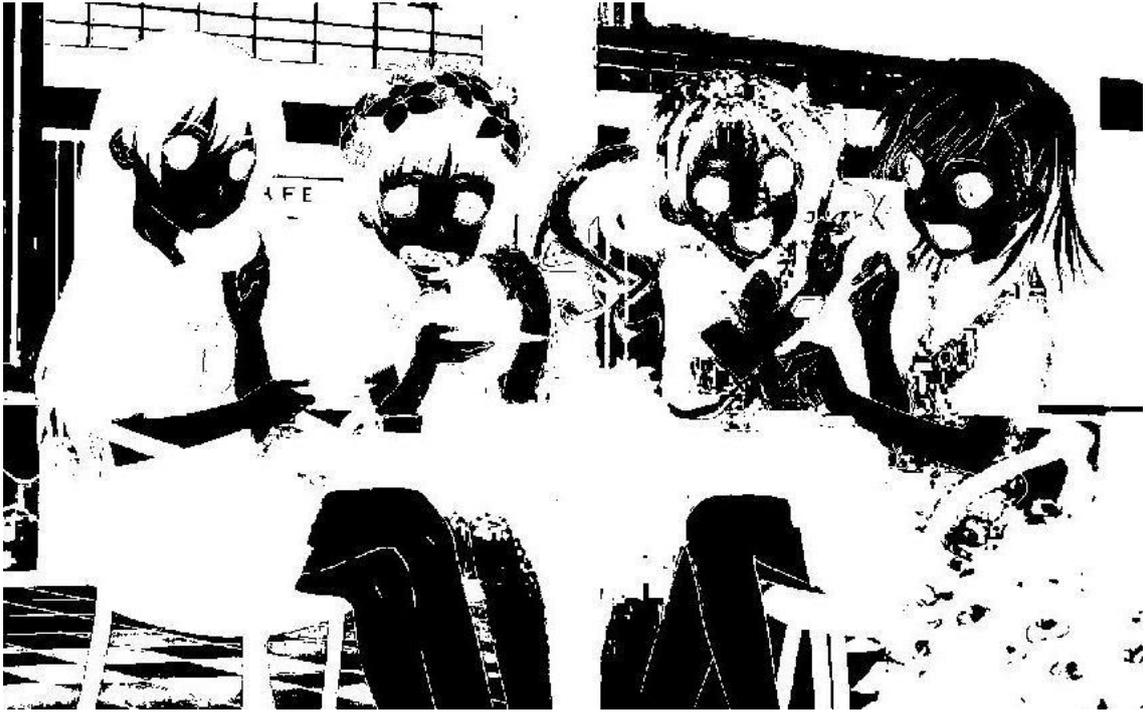

Figure 22. Skin extracted using Method-II

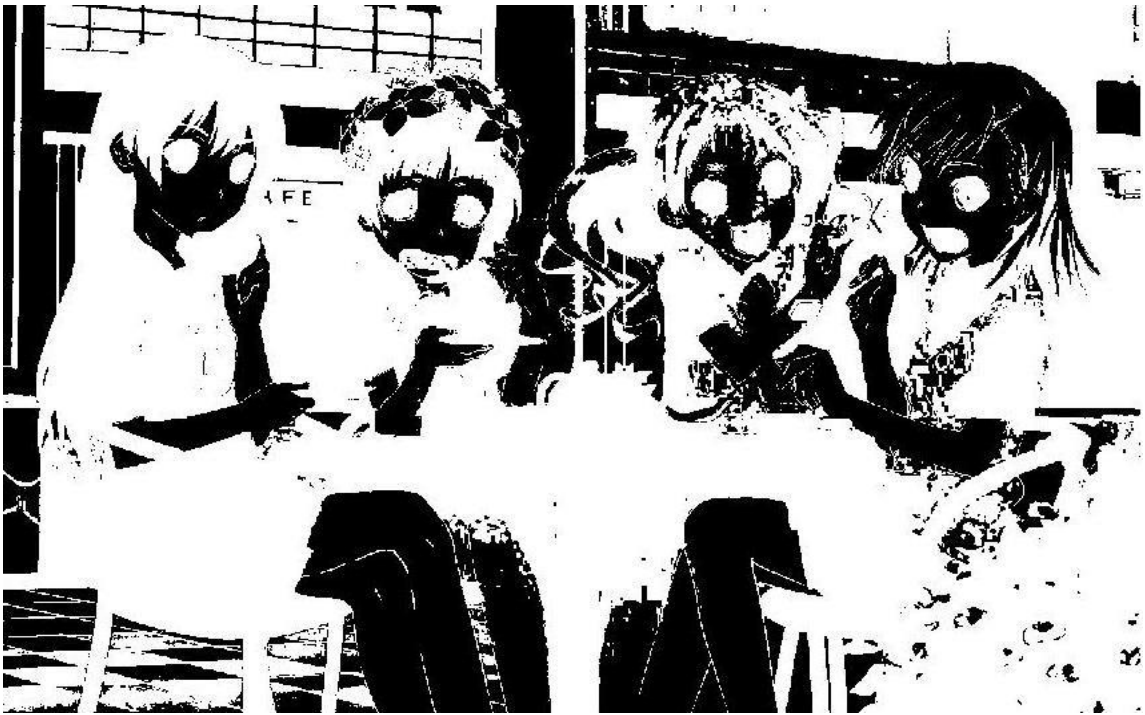

Figure 23. Skin extracted using Method-III

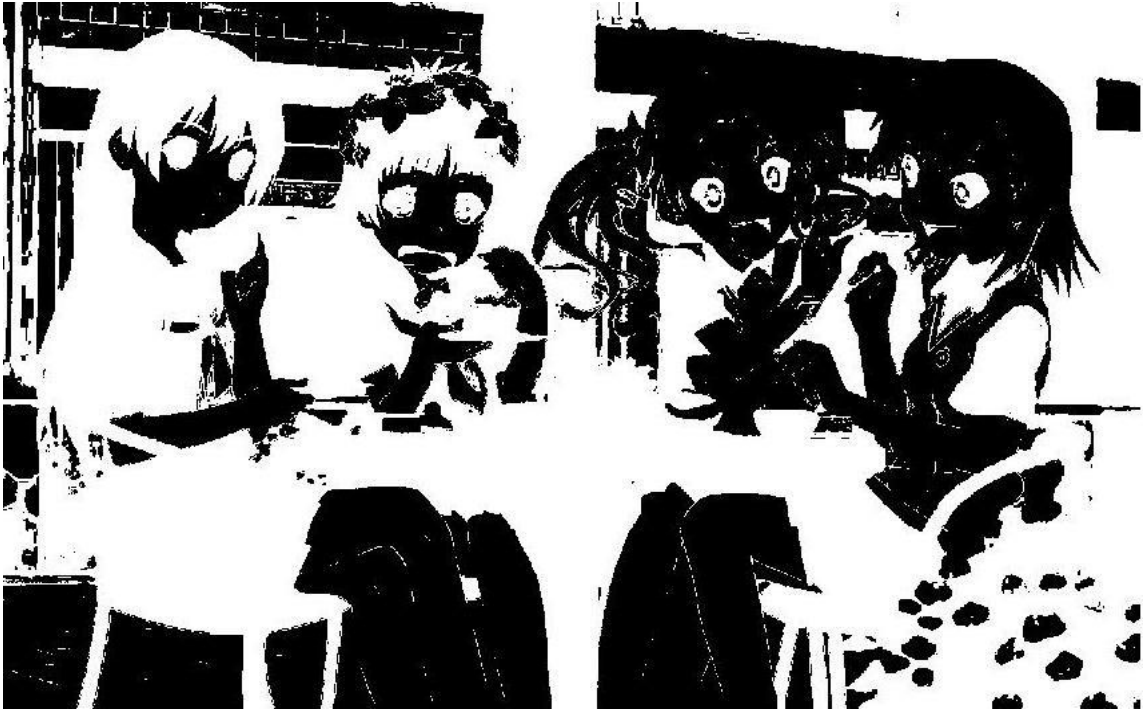

Figure 24. Skin extracted using Kovac method

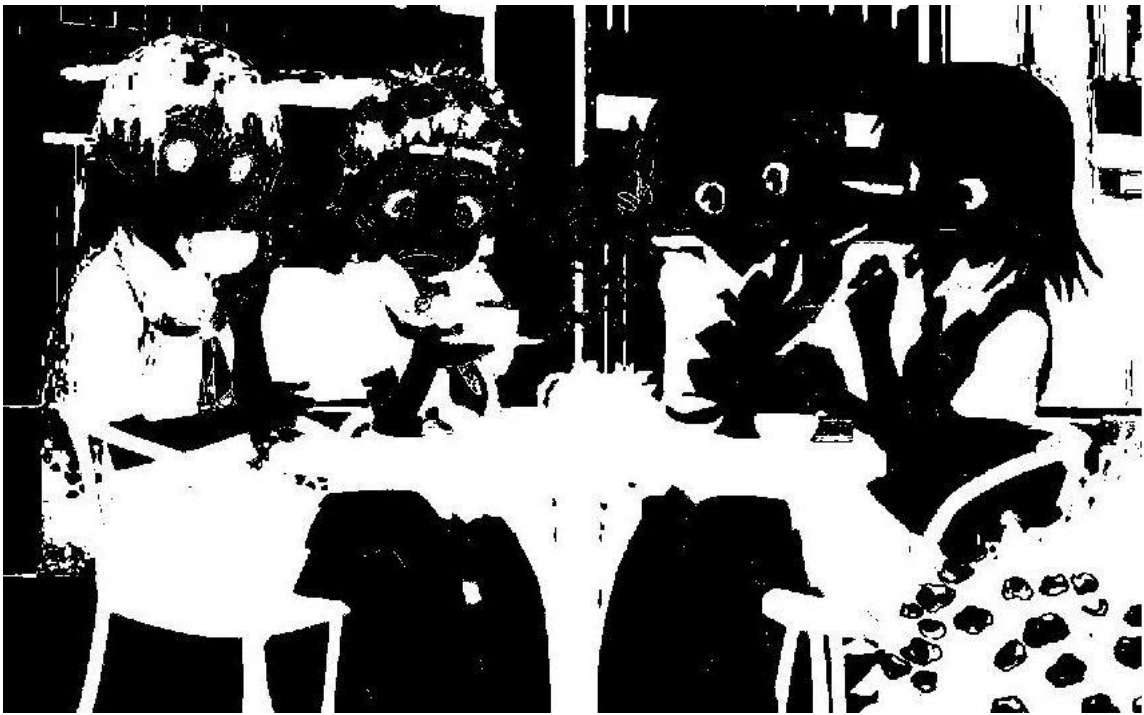

Figure 25. Skin extracted using Osman method

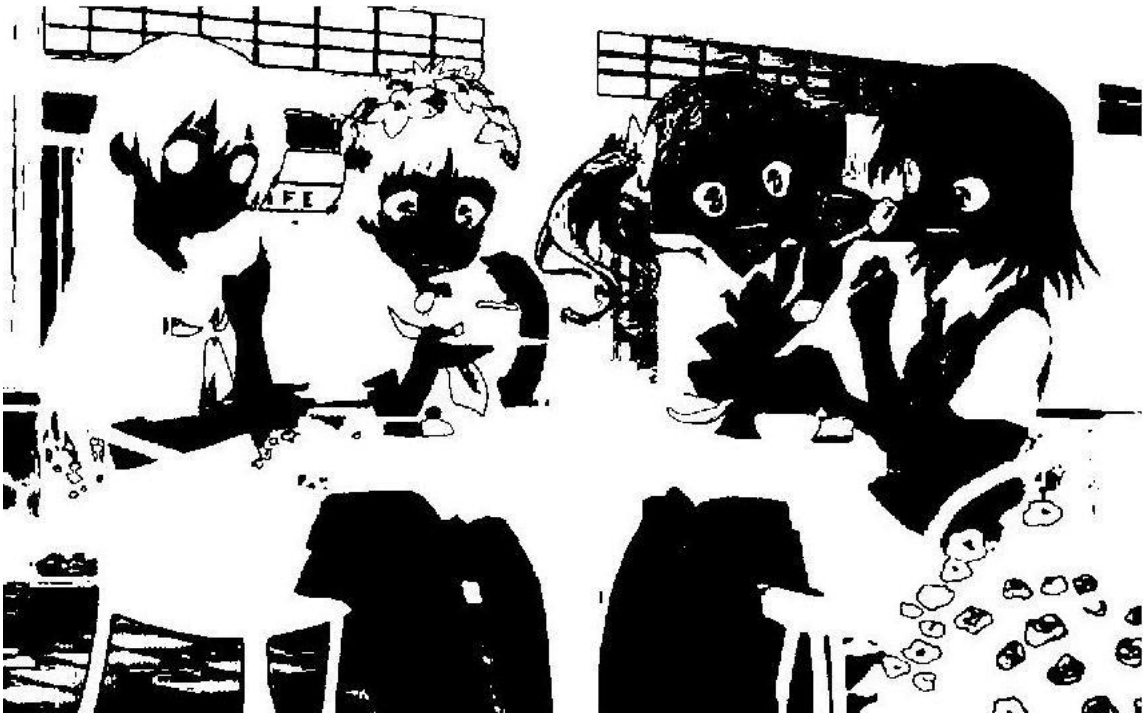

Figure 26. Skin extracted using Saleh method

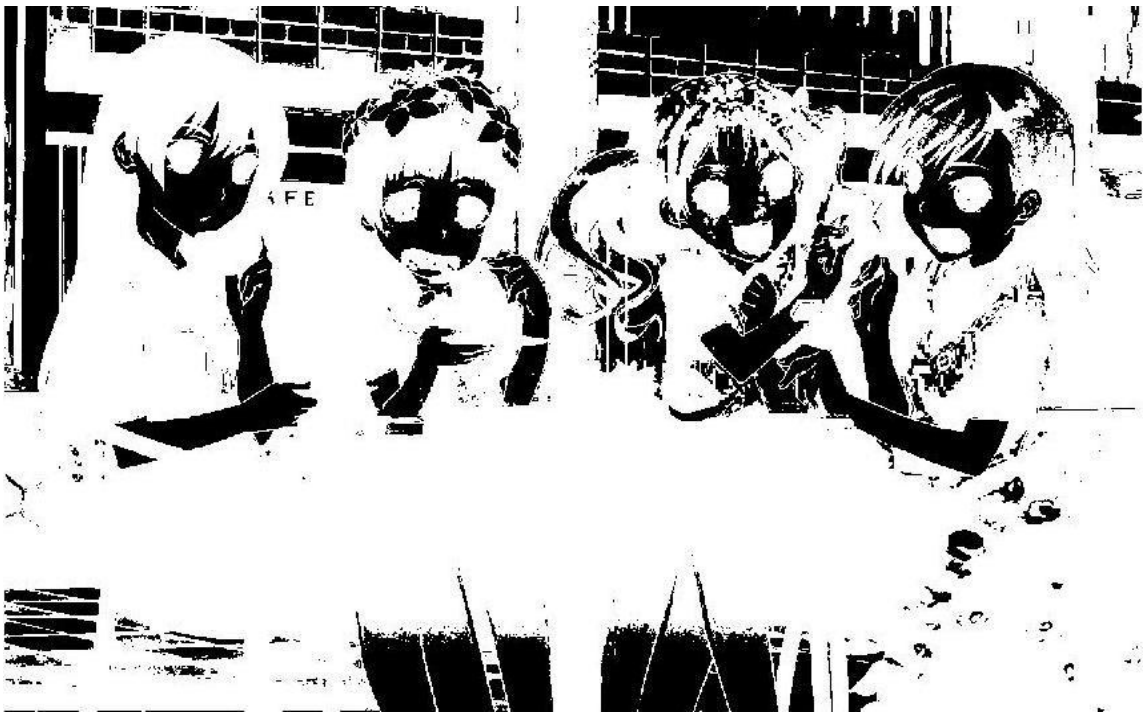

Fig. 27. Skin extracted using Takayama method

## 5. CONCLUSIONS

In this paper, we have presented three different methods to extract skin from a cartoon image. We have also showed the results of different methods in comparison with Kovac, Osman, Saleh and Takayama's methods. Our methods are based on RGB values of pixel and their comparative relations. It is evident that, only a certain range of value cannot strictly define skin in an image. We need to incorporate the shape of the character to detect the skin region. Moreover, we have also showed that, skin color can be different for different genders and races. Also, the skin color of humans and anime characters are distinctive. A method or a set of rules based on only RGB values will have limitations.

In future, we will include shape recognition of anime characters, and utilize the detected shapes along with the RGB value-based rules for skin detection. Also we will use clustering to find similarly colored pixels, so that we can set different conditions for skin for different races.

## Authors

Kazi Tanvir Ahmed Siddiqui received his B.Sc. degree from North South University. He is currently working at Bangladesh University of Engineering and Technology. His research interests include Artificial Intelligence, Machine Learning and Image Processing.

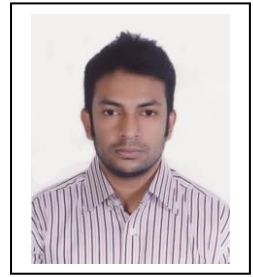

Abu Wasif is currently an Assistant Professor at Bangladesh University of Engineering and Technology. His main research interests are in Artificial Intelligence and Machine Learning.

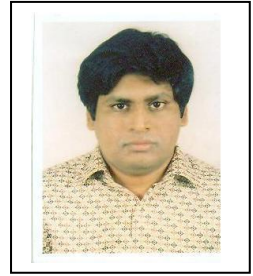